\documentclass[10pt,letterpaper]{article}

\usepackage{cogsci}

\usepackage{graphicx}

\usepackage{amsmath} 

\usepackage{booktabs}

\usepackage{enumitem}

\cogscifinalcopy % Uncomment this line for the final submission 

\usepackage{pslatex}
\usepackage{apacite}
\usepackage{float} % Roger Levy added this and changed figure/table
\title{MEKiT: Multi-source Heterogeneous Knowledge Injection Method via Instruction Tuning for Emotion-Cause Pair Extraction}
\author{Shiyi Mu, Yongkang Liu, Shi Feng$^*$,  \\
  \textbf{Xiaocui Yang, Daling Wang, Yifei Zhang} \\
        Northeastern University, Shenyang, China \\
        \texttt{mushiyi@stumail.neu.edu.cn,misonsky@163.com} \\
        \texttt{\{fengshi,yangxiaocui,wangdaling,zhangyifei\}@cse.neu.edu.cn }
}

\begin{document}

\maketitle

\begin{abstract}
% Include no author information in the initial submission, to facilitate
% blind review.  The abstract should be one paragraph, indented 1/8~inch on both sides,
% in 9~point font with single spacing. The heading ``{\bf Abstract}''
% should be 10~point, bold, centered, with one line of space below
% it. This one-paragraph abstract section is required only for standard
% six page proceedings papers. Following the abstract should be a blank
% line, followed by the header ``{\bf Keywords:}'' and a list of
% descriptive keywords separated by semicolons, all in 9~point font, as
% shown below.
Although large language models (LLMs) excel in text comprehension and generation, their performance on the Emotion-Cause Pair Extraction (ECPE) task, which requires reasoning ability, is often underperform smaller language model. The main reason is the lack of auxiliary knowledge, which limits LLMs' ability to effectively perceive emotions and reason causes. To address this issue, we propose a novel \textbf{M}ulti-source h\textbf{E}terogeneous \textbf{K}nowledge \textbf{i}njection me\textbf{T}hod, MEKiT, which integrates heterogeneous internal emotional knowledge and external causal knowledge. Specifically, for these two distinct aspects and structures of knowledge, we apply the approaches of incorporating instruction templates and mixing data for instruction-tuning, which respectively facilitate LLMs in more comprehensively identifying emotion and accurately reasoning causes. Experimental results demonstrate that MEKiT provides a more effective and adaptable solution for the ECPE task, exhibiting an absolute performance advantage over compared baselines and dramatically improving the performance of LLMs on the ECPE task.

\end{abstract}

\section{Introduction}

Emotion is fundamental to human cognition and behavior, serving as a crucial role in decision-making, social interactions, and communication processes~\cite{ekman1992argument}. Emotions not only reflect an individual's internal psychological state but also regulate adaptive behaviors, such as coping with stress, building relationships, and responding to environmental stimuli~\cite{gross1998emerging}. Identifying the causes behind emotions is central to emotion regulation theories~\cite{gross2007emotion}. According to the emotion cognitive appraisal theory~\cite{lazarus1991emotion}, emotions arise from an individual's evaluation of event or situation, emphasizing the importance of understanding not only the emotions themselves but also the latent triggers. The Emotion-Cause Pair Extraction (ECPE) task~\cite{xia2019emotion} is essential for advancing the theory of emotion cognition.

ECPE involves extracting emotion-cause pairs from a given document with multiple clauses.
% ECPE is to extract emotion-cause pairs from the given document, where the given document contains several clauses.
% The clauses expressing emotions are referred to as “emotion clause” and clauses causing emotions are referred to as “cause clauses”.
The existing ECPE methods can be categorized into pipeline extraction~\cite{xia2019emotion} and joint extraction~\cite{wei2020effective,song2020end,li2023effective}.
% Early research primarily concentrated on the two-step framework and the end-to-end framework \cite{wei2020effective,song2020end,li2023effective}. 
Pipeline extraction methods extract emotion and cause separately, followed by emotion-cause pairing and filtering~\cite{xia2019emotion}, but they fail to adequately capture the interdependence between emotion and cause clauses.
% The two-step framework first performs individual emotion extraction and cause extraction by using two classifiers and then performs emotion cause pairing and filtering. 
% However, this approach does not adequately capture the interdependence between emotion and cause clauses.
The joint extraction methods can simultaneously identify emotions and causes, directly establishing the association between them during extraction process, thereby capturing their inherent dependency~\cite{wei2020effective,song2020end,li2023effective}. 
% The end-to-end framework directly extracts emotion clauses and their corresponding cause clauses from the text through joint training, thereby avoiding the independence issue between emotion and cause extraction present in traditional multi-stage approaches.
Although these approaches have achieved promising results, their excessive reliance on positional information and limitations in effectively capturing the latent semantic information of causal clauses still constrain their reasoning ability.
% Although these approaches have achieved promising results, they exhibited limited reasoning capabilities, an excessive reliance on positional information, and insufffcient ability to effectively capture the latent semantic information of causal clauses~\cite{ding2020experimental,zheng2022ueca}. 

\begin{figure}[t]
  \includegraphics[width=\columnwidth]{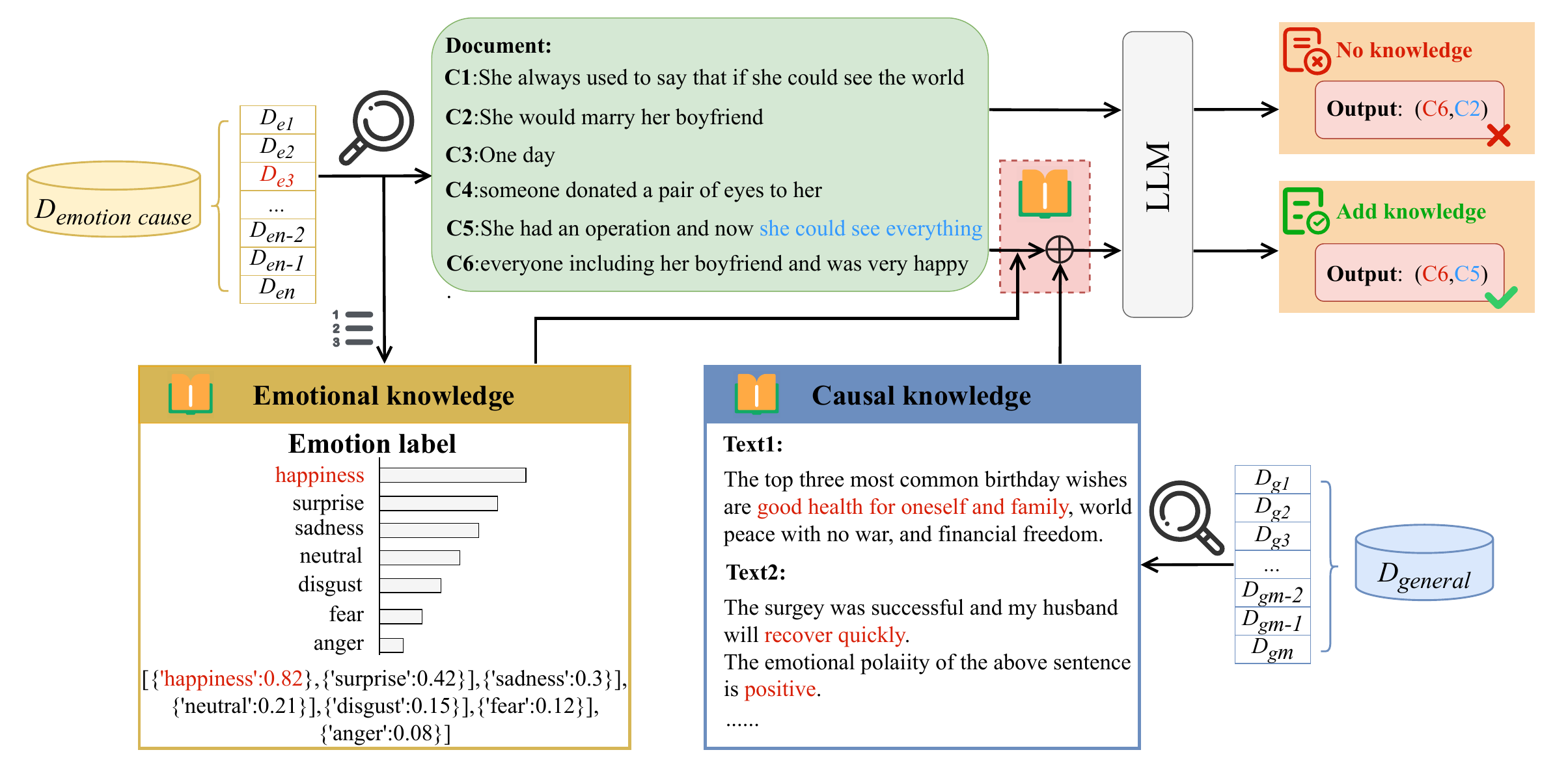}
  \caption{An example of ECPE injected with emotional knowledge and causal knowledge.}
  \label{fig:example}
\end{figure}

Insufficient reasoning ability problems still exist in the era of large language models (LLMs). Due to the lack of auxiliary knowledge, LLMs perform poorly on reasoning ability based on inherent explicit semantic understanding capabilities and brief prompt templates, facing challenges in understanding the nuanced relationships between emotions and corresponding causes~\cite{wang2023chatgpt,wu2024enhancing}. Some researchers have attempted to introduce external knowledge to alleviate this problem in similar tasks~\cite{zhao2023knowledge,li2021past,li2021enhancing}. However, most of these methods focus solely on single-source emotional knowledge, ignoring the deep relational knowledge between emotion and cause. As a result, these methods improve emotion perception ability in a one-sided manner, but it lacks the ability to accurately attribute emotions, preventing precise extraction of emotion-cause pairs. As shown in Figure~\ref{fig:example}, emotional knowledge only provides the emotional tendencies of this example (i.e., \textit{happiness $\textgreater$ surprise $\textgreater$ ... $\textgreater$ anger}), without enhancing the model's ability to reason causes effectively. In this case, emotional knowledge enhancement instills confidence in LLM regarding its own judgments and LLM can recognize that the sentiment of the document is \textbf{happiness}, but it cannot accurately trace the cause of that happiness (i.e., \textit{...she could see everything...}). Improving emotional reasoning ability may require different knowledge, and the structure of these knowledge is generally different. The integration of such multifaceted and heterogeneous knowledge presents a significant challenge.

To address these problems, we propose a \textbf{M}ulti-source h\textbf{E}terogeneous \textbf{K}nowledge \textbf{i}njection me\textbf{T}hod (MEKiT) based on instruction-tuning, which offers a unified method that simultaneously enhances the model's ability to perceive emotions and reason causes. MEKiT conducts instruction-tuning by incorporating emotional knowledge into instruction templates and mixing causal knowledge into the emotion cause dataset. For each document, it provides structured label distribution knowledge associated with the corresponding emotional label. In cases where label distribution knowledge is unavailable, the method offers coarse-grained emotional polarity knowledge instead. Both label distribution and polarity knowledge effectively enhance emotion perception. To enhance the model's ability to reason causes, MEKiT introduces causal knowledge in natural text. As shown in Figure~\ref{fig:example}, the \textit{text1} “\textit{The top three most common birthday wishes are good health for oneself and family...}” typically reflects positive emotions (birthday wishes are generally positive) and corresponding reason (family members' good health). The \textit{text2} “\textit{The surgey was successful and my husband will recover quickly...is positive}” conveys \textbf{positive tendency} (\textbf{happiness}) and the corresponding reason (the family recovered well). By mixing auxiliary causal knowledge during instruction-tuning, the model’s ability to reason causes is enhanced. 
In summary, we make the following contributions: 
\setenumerate[1]{itemsep=0pt,partopsep=0pt,parsep=\parskip,topsep=0pt}
\setitemize[1]{itemsep=0pt,partopsep=0pt,parsep=\parskip,topsep=0pt}
\setdescription{itemsep=0pt,partopsep=0pt,parsep=\parskip,topsep=0pt}
\begin{itemize}[leftmargin=*]
    \item We introduce a novel multi-source heterogeneous knowledge injection method leveraging instruction-tuning, which is subsequently implemented in LLMs to enhance their reasoning capabilities. 
    \item We construct instruction templates to better incorporate emotional knowledge and explore the optimal causal knowledge mixing ratio that maximizes the performance of the method in the instruction-tuning stage.
    \item Experimental results over various settings demonstrate that our method not only achieves outstanding performance but also shows strong generality on the ECPE task, as demonstrated by its effective compatibility with different LLMs.
\end{itemize}

\section{Related Work}

% First level headings should be in 12~point, initial caps, bold and
% centered. Leave one line space above the heading and 1/4~line space
% below the heading.

\subsection{Emotion-Cause Pair Extraction Task}

Emotion-Cause Pair Extraction (ECPE) and a two-step method are first proposed \cite{xia2019emotion}. However, this method will suffer from cascading errors and fail to adequately capture the interdependence between emotion and cause clauses. Subsequently, many end-to-end approaches are proposed and achieve better results \cite{chen2020end,song2020end,ding2020end,li2023effective}. Although most existing approaches differ in model structures, they introduce relative position information, which is limited by the position bias. Therefore, some researchers design a BERT-based prompt learning joint extraction method which realizes the extraction of multiple sub-tasks by constructing special prompts to solve the ECPE task and significantly improving performance on the ECPE task \cite{zheng2022ueca}. However, this approach still struggles to effectively capture the latent semantic information of cause clauses.

\subsection{Knowledge Injection in Sentiment Analysis}

External knowledge can include sentiment lexicons, knowledge graphs, or other knowledge sources, which help models better understand contextual semantics and emotional triggering mechanisms. KBCIN utilized event-centered commonsense knowledge to provide explicit causal clues~\cite{zhao2023knowledge}. Commonsense knowledge is also used to enrich the edges of the graph and enhance psychological interactions modeling between utterances~\cite{li2021past}. These methods based on single-source emotional knowledge injection often bring information overlap and cannot cover much effective information. In open-domain dialogue generation, researchers integrate different types of knowledge and improve knowledge coverage to generate richer and more informative dialogue responses~\cite{wu2021more}. A cognitive stimulation dialogue system integrates external knowledge sources like the Chinese EmoBank~\cite{lee2022chinese} to calculate weight values to words, enabling the model to prioritize high-value terms. The designed multi-source interaction mechanism further combines emotional support strategies with cognitive stimulation therapies to generate responses that promote mental health ~\cite{jiang2023cognitive}.

% \subsubsection{Third Level Headings}

% Third level headings should be 10~point, initial caps, bold, and flush
% left. Leave one line space above the heading, but no space after the
% heading.

\begin{figure*}[t]
  \includegraphics[width=1.0\linewidth]{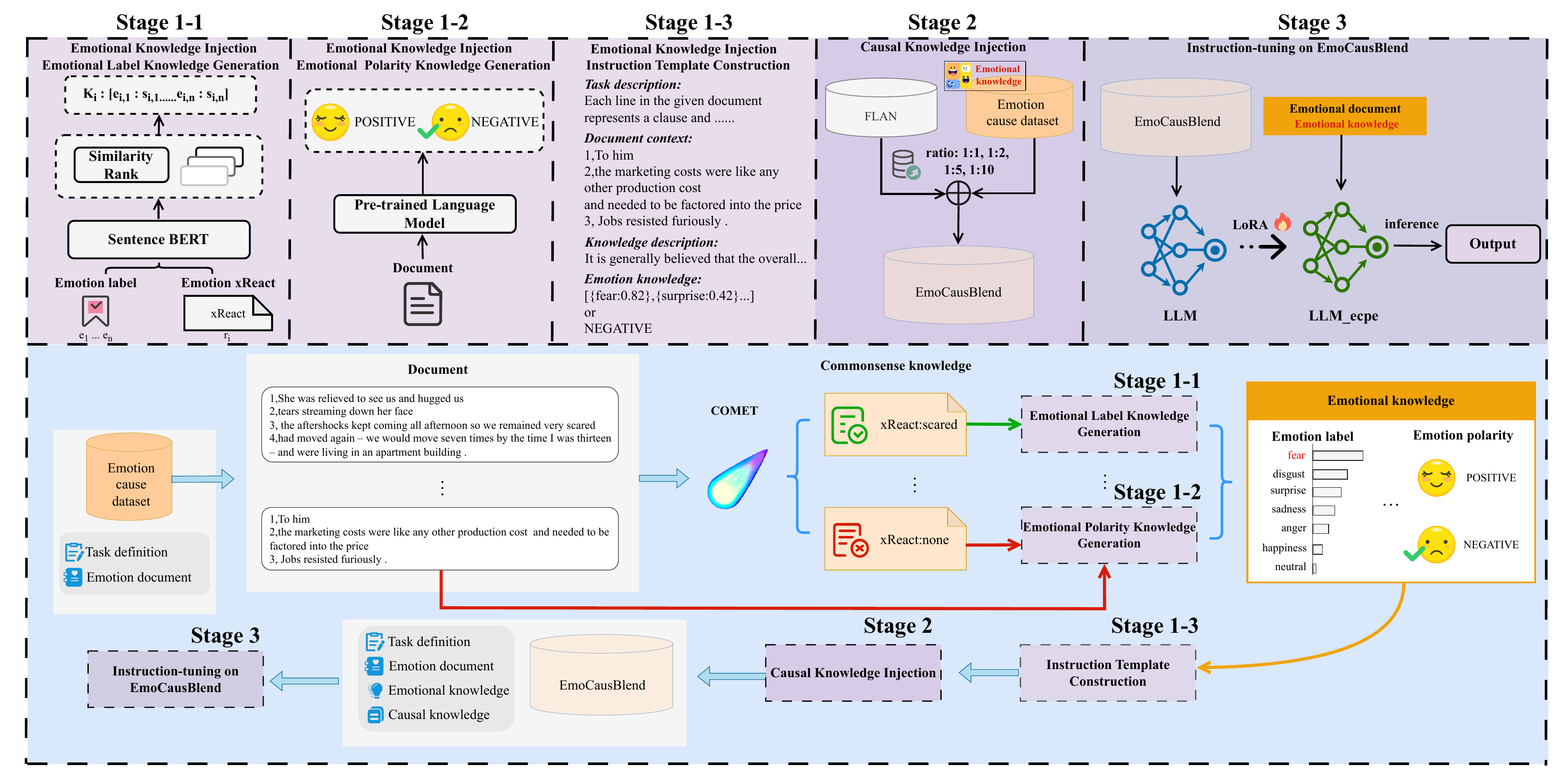}
  \caption {Overview of MEKiT.}
\label{fig:framework}
\end{figure*}

\section{Method}

In this section, we introduce a multi-source heterogeneous knowledge injection method, MEKiT, which consists of three main stages, \textit{emotional knowledge injection}, \textit{causal knowledge injection}, \textit{instruction-tuning on EmoCausBlend}. Our workflow is illustrated in Figure~\ref{fig:framework}. Stage 1 generates textual emotional knowledge for each document, which encompasses either emotional label knowledge or emotional polarity knowledge and this emotional knowledge is employed to design instruction templates, forming an emotion cause dataset enriched with emotional knowledge. In the Stage 2, we incorporate open-source causal data into the dataset augmented with emotional knowledge from Stage 1 to form the final training set named \textit{EmoCausBlend}. In the Stage 3, the \textit{EmoCausBlend} is utilized to fine-tune LLMs. According to emotion cognitive appraisal theory, emotions originate from individuals' appraisal of events, which is an entirely implicit inducement. These stages are dedicated to promoting LLM to understand subtle and potential emotions and explore the causes behind them through instruction-tuning.

% \begin{figure*}[t]
%   \includegraphics[width=1.0\linewidth]{img/framework5.pdf}
%   \caption {Overview of our proposed MEDIA.}
% \label{fig:fig5}
% \end{figure*}

\subsection{Emotional Knowledge Injection}

This stage consists of three sub-stages: emotional label knowledge generation, emotional polarity knowledge generation, and instruction template construction. In summary, our method generates emotional knowledge based on the original dataset (Stage 1-1, Stage 1-2) and inject it into instruction templates (Stage 1-3).

COMET \cite{bosselut2019comet} is an excellent framework designed to generate high-quality commonsense knowledge. The xReact relation in COMET represents the emotion experienced by an individual in an event, closely tied to the emotional tone of the most intense clause in the document. In this stage, we first apply the COMET BART version trained on $ \text{ATOMIC}_{20}^{20} $ \cite{hwang2021symbolic} as the intermediate knowledge generation tool. The tool outputs the potential emotional reaction of an individual to the given scenario, which can serve as intermediate emotional knowledge. Given a document context \textit{D\textsubscript{i}}, the commonsense result is as follows: 
\begin{equation}
\begin{aligned}
r_i &= \text{COMET}_{\text{BART}}(D_i,R)
\end{aligned} \tag{1}
\end{equation}
where \textit{R} = xReact. The content of \textit{r\textsubscript{i}} can be categorized into two types: “\textit{non-none}” and “\textit{none}”, where “\textit{non-none}” is used for emotional label knowledge generation and “\textit{none}” for emotional polarity knowledge generation.
% The \textit{non-none} category represents valid knowledge with a potential emotional reaction. In contrast, the \textit{none} category represents invalid knowledge, indicating that the tool considers the input scenario insufficient to evoke a significant emotional response.
% where the tool outputs the potential emotional reaction of an individual to the given scenario, which can serve as intermediate emotional knowledge, providing a basis for subsequent steps.

\subsubsection{Emotional Label Knowledge Generation}

For the \textit{non-none} xReact relation, we employ the bidirectional encoder SBERT \cite{reimers2019sentence} to identify the most semantically similar emotional label (fear, disgust, sadness, happiness, surprise, anger, neutral) based on cosine similarity. As shown in Stage 1-1 of Figure~\ref{fig:framework}, SBERT generates independent CLS embeddings for the target xReact \textit{r\textsubscript{i}} and each label \textit{e\textsubscript{j}} in emotional category, which encompasses \textit{n} types of emotional labels. After calculating the cosine similarity, each emotional label and its corresponding score are sorted from high to low to form a distribution list and incorporated into the instruction template, which aims to provide the model with enriched emotional context and supplementary information that may be challenging for the model to infer independently. Cosine similarity score $s_{i,j}$ between \textit{r\textsubscript{i}} and \textit{e\textsubscript{j}} is written as: 
\begin{equation}
\begin{aligned}
s_{i,j} = \text{Sim}(\text{SBERT}(r_i, e_j))
\end{aligned} \tag{2}
\end{equation}

\subsubsection{Emotional Polarity Knowledge Generation}

As shown in Stage 1-2 of Figure~\ref{fig:framework}, for the \textit{none} xReact relation, the intermediate knowledge generated by COMET is deemed invalid, accounting for 43\% of the intermediate results in emotion cause dataset, based on statistical analysis. In such instances, we use emotional polarity knowledge (POSITIVE , NEGATIVE) derived from Hugging Face's pipeline API, a pre-trained Transformer-based language model indicated by \textit{pipeline\_classifier} as an auxiliary emotional knowledge. Since emotional polarity is coarse-grained, it achieves higher classification accuracy, minimizing noise and effectively guiding the model's reasoning process. 

The complete formula of overall emotional knowledge expression \textit{K\textsubscript{i}} is as follows:
\begin{equation}
K_i = 
\begin{cases} 
\{(e_{j}, s_{i,j}) \mid j = 1, 2, \ldots, n\} & \text{if } r_i \neq \text{None}, \\
pipeline\_classifier(D_i) & \text{if } r_i = \text{None},
\end{cases}\tag{3}
\end{equation}

\subsubsection{Instruction Template Construction}

In the task of emotion cause analysis, designing appropriate instruction templates is crucial for improving model performance. This approach more effectively guides LLMs to uncover potential causes, thereby enhancing model accuracy and controllability. As shown in the prompt template in Stage 1-3 in Figure~\ref{fig:framework}, our instruction templates are designed to include four essential elements: a concise and clear task description, document context, emotional knowledge description and emotional knowledge obtained from the previous sub-stages. These elements are carefully structured to provide contextual clarity and logical coherence, enabling the model to better understand the relationships between emotions and underlying causes. After this stage, we obtain the emotion cause dataset injected with emotional knowledge based on the original training set. 

\subsection{Causal Knowledge Injection}
In the context of injecting emotional knowledge into instruction templates, we utilize a large-scale, open-source English general dataset FLAN~\cite{wei2021finetuned} containing over a million data, covering 60 diverse NLP tasks, such as common sense reasoning, sentiment analysis. Based on the predefined mixing ratio, we extract causal knowledge in the form of natural text from FLAN by calculating the similarity between each entry in the emotion cause dataset and the FLAN corpus. As shown in Stage 2 in Figure~\ref{fig:framework}, since the sources and forms of emotional knowledge and causal knowledge are different, unlike the process of injecting emotional knowledge, causal data is mixed into the emotion cause dataset enriched with emotional knowledge from Stage 1 to construct EmotionCausalBlend (\textit{EmoCausBlend}) training set for the instruction-tuning stage (Stage 3). The introduction of causal knowledge aims to increase prior knowledge and experience, improve cognitive abilities as well as constantly adjust the mixed ratio of original emotion cause data and causal data. By doing so, we not only enhance the model's overall reasoning and language understanding capabilities but also ensure that it can effectively balance emotional sensitivity with broader contextual comprehension.

\subsection{Instruction-Tuning on \textit{EmoCausBlend}}
As shown in Figure~\ref{fig:framework}, in this phase, we perform instruction-tuning on LLM using the \textit{EmoCausBlend} training set, which is injected multi-source knowledge and constructed in the previous stages. To achieve efficient and parameter-efficient fine-tuning, we apply LoRA-Tuning \cite{hu2021lora}, a low-rank adaptation method that allows for targeted updates to specific model layers without requiring a full retraining of LLMs. According to the original training method of LLM, we adopt the next token prediction loss as the objective function to quantify the discrepancy between the model's predicted outputs and the ground-truth tokens. Therefore, the loss calculation is denoted as \textit{L} and is defined as follows:
% \begin{equation}
% \mathcal{L} = \sum_{i=1}^{N} -\log P(y_{i} \mid x_{i}, \theta) \tag{3}
% \end{equation}
\begin{equation}
L = \sum_{i=1}^{N} -\log P(y_{i} \mid x_{i}, \theta) \tag{4}
\end{equation}
where \textit{y\textsubscript{i}} represents the token of the corresponding emotion-cause pairs for the given task input sample \textit{x\textsubscript{i}}. \textit{N} stands for the total number of documents in the dataset, while \(\theta \) represents the parameters of the LLM. Since we focus on the model’s capabilities in the ECPE task, we restrict the case to extracting emotion-cause pairs from the test set enriched with emotional knowledge during the inference phase for consistency.

% \begin{table}[H]
% \begin{center} 
% \caption{Sample table title.} 
% \label{sample-table} 
% \vskip 0.12in
% \begin{tabular}{ll} 
% \hline
% Error type    &  Example \\
% \hline
% Take smaller        &   63 - 44 = 21 \\
% Always borrow~~~~   &   96 - 42 = 34 \\
% 0 - N = N           &   70 - 47 = 37 \\
% 0 - N = 0           &   70 - 47 = 30 \\
% \hline
% \end{tabular} 
% \end{center} 
% \end{table}

% \subsection{Figures}

% All artwork must be very dark for purposes of reproduction and should
% not be hand drawn. Number figures sequentially, placing the figure
% number and caption, in 10~point, after the figure with one line space
% above the caption and one line space below it, as in
% Figure~\ref{sample-figure}. If necessary, leave extra white space at
% the bottom of the page to avoid splitting the figure and figure
% caption. You may float figures to the top or bottom of a column, and
% you may set wide figures across both columns.

% \begin{figure}[H]
% \begin{center}
% \fbox{CoGNiTiVe ScIeNcE}
% \end{center}
% \caption{This is a figure.} 
% \label{sample-figure}
% \end{figure}

\section{Experiment}

\subsection{Dataset and Metrics}

We selected English dataset NTCIR-13 Workshop \cite{gao2017overview}, composed of English novels. We then evaluate our results using precision (P), recall (R), and F1-score metrics, as used in the past work on ECPE task \cite{xia2019emotion}. 

\subsection{Baselines}

For this dataset, we compare the proposed method with Indep, E2E-PExtE~\cite{singh2021end}, ECPE-2D ~\cite{ding2020ecpe}, ECPE-MLL~\cite{ding2020end}, and IA-ECPE~\cite{huang2023emotion}. In addition, we evaluate ChatGPT series LLMs in few-shot scenario, including GPT-3.5-turbo and GPT-4o. We also evaluate the performance of some open source trainable scalable mainstream LLMs in few-shot scenario and instruction-tuning scenario, including Vicuna-7B~\cite{chiang2023vicuna}, LLaMA2-7B~\cite{touvron2023llama}, LLaMA3-8B-Instruct~\cite{touvron2023llama}, Qwen2.5-7B-Instruct~\cite{yang2024qwen2}, Gemma-2-9B-it~\cite{team2024gemma}.

% \begin{table}[H]
% \begin{center} 
% \caption{Comparison results of baselines and our method. The first experimental group represents the results of several existing methods, the second group denotes instruction-tuning method, and the third group reflects few-shot learning methods.} 
% \label{sample-table} 
% \vskip 0.12in
% \begin{tabular}{l c c c} 
% \hline
% Method    &  P(\%)    &  R(\%)    &  F1(\%) \\
% \hline
% Indep       &   46.94    &   41.02   &   43.67\\
% ECPE-2D       &   60.49    &   43.84   &   50.73\\
% ECPE-MLL       &   59.26    &   45.30   &   51.21\\
% E2E-PExtE       &   51.34    &   49.29   &   50.17\\
% IA-ECPE       &   60.14    &   43.03   &   50.05\\
% \midrule
% Vicuna-7B        &   55.63    &   49.53   &   52.41\\
% LLaMA2-7B           &   51.06    &   45.46   &   48.09\\
% LLaMA3-8B-Instruct      &   60.92    &   54.23   &   57.38\\
% Qwen2.5-7B-Instruct      &   58.39    &   52.35   &   55.21\\
% Gemma-2-9B-it      &   61.05    &  54.55   &   57.62\\
% \midrule
% Gemma-2-9B-it (1-shot)       &   15.84    &   15.00   &   15.41\\
% GPT-3.5-turbo (1-shot)       &   17.48    &   15.67   &   16.53\\
% GPT-4o (1-shot)       &   28.52    &   25.39   &   26.87\\
% \midrule
% MEKiT         &   \textbf{64.88}    &   \textbf{60.82}   &   \textbf{62.78}\\
% \hline
% \end{tabular} 
% \end{center} 
% \end{table}

\subsection{Results and Analysis}

\subsubsection{ECPE with Instruction-Tuning}

Table~\ref{table1} shows our evaluation of the performance of the most popular LLMs (GPT-3.5-turbo, GPT-4o) and open source trainable LLMs in instruction-tuning and few-shot scenarios. Although ChatGPT has remarkable competitive performance in general tasks \cite{sun2023pushing,qin2023chatgpt}, the instruction-tuning method outperforms GPT-3.5-turbo and GPT-4o in few-shot settings. In the absence of knowledge injection, Gemma-2-9B-it based on instruction-tuning can outperform other fine-tuning-based works such as ECPE-MLL based on multi-label learning, and IA-ECPE based on interactive attention on the ECPE task, improving the recall by 9.25\%, and 11.52\%, respectively. Notably, Gemma-2-9B-it achieves the best results among trainable LLMs via instruction-tuning. Therefore, we select Gemma-2-9B-it as the backbone for subsequent experiments. However, despite the improvements in recall, the precision advantage of instruction-tuned models remains less significant. This suggests that while instruction-tuned LLMs exhibit strong representational capacity and learning abilities, these models still misclassify more negative pairs as positive, thereby improving recall, but this relaxed strategy also leads to a decrease in precision due to the lack of knowledge crucial for precise emotion cause prediction. As shown in the last row of Table~\ref{table1}, the result of our Gemma-2-9B-it-based approach MEKiT on the \textit{EmoCausBlend}, which outperforms other baselines and proves the importance of knowledge injection.

\begin{table}[H]
\begin{center} 
\caption{Performance comparison of MEKiT and other baselines. The symbol * indicates the results following the process of instruction-tuning.} 
\label{table1} 
\vskip 0.12in
\begin{tabular}{l c c c} 
\hline
Method    &  P(\%)    &  R(\%)    &  F1(\%) \\
\hline
Indep       &   46.94    &   41.02   &   43.67\\
ECPE-2D       &   60.49    &   43.84   &   50.73\\
ECPE-MLL       &   59.26    &   45.30   &   51.21\\
E2E-PExtE       &   51.34    &   49.29   &   50.17\\
IA-ECPE       &   60.14    &   43.03   &   50.05\\
\midrule
Vicuna-7B\textsuperscript{*}        &   55.63    &   49.53   &   52.41\\
LLaMA2-7B\textsuperscript{*}           &   51.06    &   45.46   &   48.09\\
LLaMA3-8B-Instruct\textsuperscript{*}      &   60.92    &   54.23   &   57.38\\
Qwen2.5-7B-Instruct\textsuperscript{*}      &   58.39    &   52.35   &   55.21\\
Gemma-2-9B-it\textsuperscript{*}      &   61.05    &  54.55   &   57.62\\
\midrule
Gemma-2-9B-it (1-shot)       &   15.84    &   15.00   &   15.41\\
GPT-3.5-turbo (1-shot)       &   17.48    &   15.67   &   16.53\\
GPT-4o (1-shot)       &   28.52    &   25.39   &   26.87\\
\midrule
MEKiT         &   \textbf{65.04}    &   \textbf{58.31}   &   \textbf{61.49}\\
\hline
\end{tabular} 
\end{center} 
\end{table}

\subsubsection{ECPE with Knowledge Injection}

Table~\ref{table2} shows the experimental results of injecting multi-source heterogeneous knowledge to the Gemma-2-9B-it and explores the best causal data mixing ratio. Two experimental groups are conducted, where ratio represents the mixed proportion of emotion cause data (ECPE task data) and causal data. Experimental results demonstrate that in the experimental group without the injection of emotional knowledge, the model achieves its highest performance at the 1:5 mixing ratio. In our method MEKiT, which incorporates emotional and causal knowledge, the model also achieves the best performance at the 1:5 mixing ratio, outperforming all baseline models across all three metrics. The multi-source knowledge injection approach we propose outperforms the LLM backbone with instruction-tuning by 3.99\%, 3.76\%, and 3.87\% in terms of P, R and F1 score, respectively. This method can effectively improve the performance of the model on this task, which improves the model’s ability to infer the causes of emotions. However, based on the results from each ratio, adding too much causal knowledge does not always lead to better results. Model fits the mapping relationship between prompts and responses in the entire dataset, the optimization objective of the model may shift toward other non-ECPE tasks if excessive injection, consequently leading to a degradation in performance on the ECPE task. Without emotional knowledge integration, moderately increasing the proportion of causal knowledge can improve model performance. This phenomenon is likely due to the model's reliance on external knowledge to enhance its reasoning ability and experiential understanding when emotional knowledge is lacking. After integrating emotional knowledge, the model acquires more emotion-related knowledge, thus injecting causal knowledge at a 1:10 ratio subsequently leads to a significant decline in performance. Incorporating moderate causal knowledge helps prevent the model from overfitting to emotion-specific data, thereby maintaining a balanced learning process.

\begin{figure}[t]
  \includegraphics[width=\columnwidth]{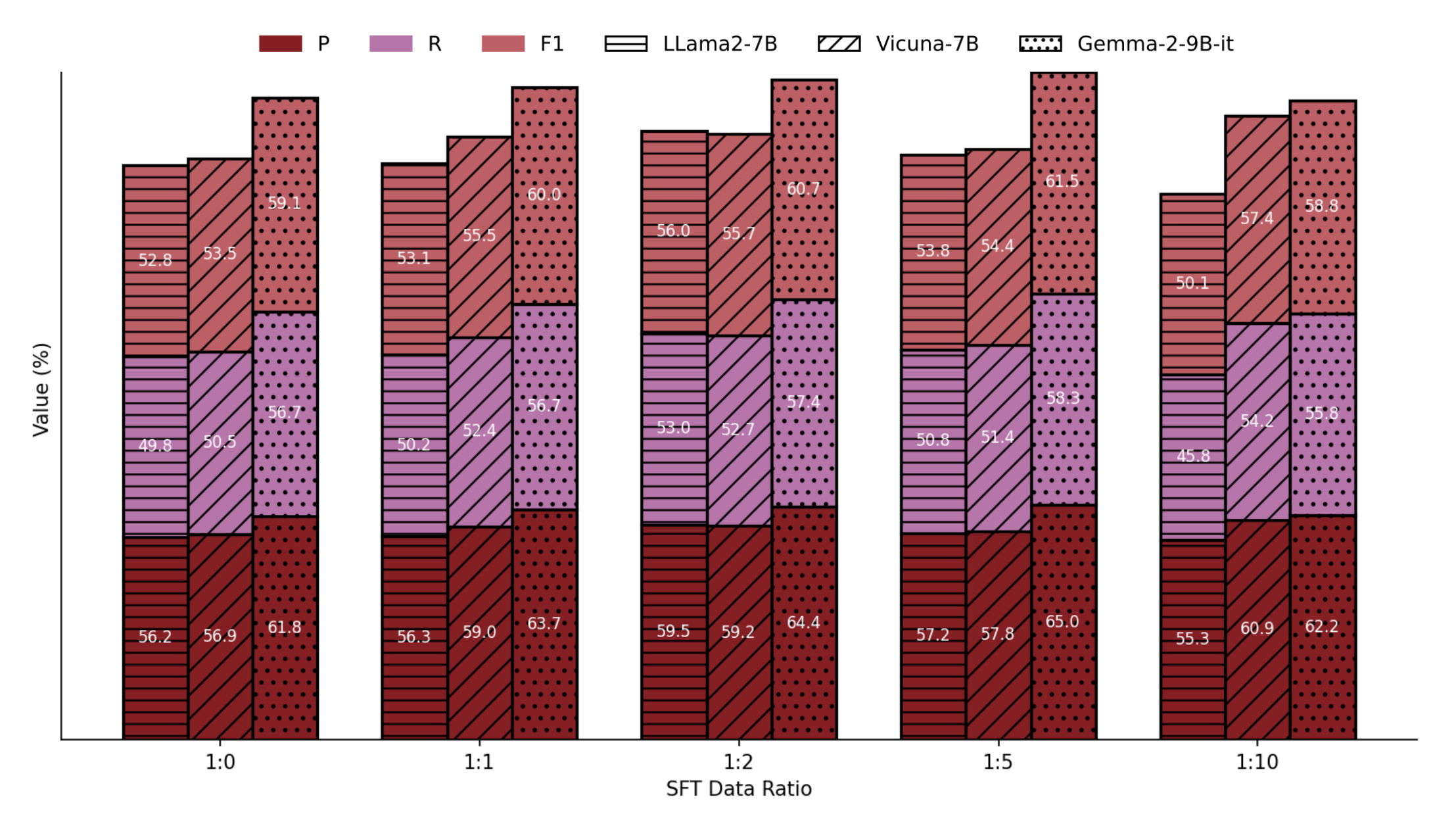}
  \caption{Results of different LLMs based on MEKiT.}
  \label{fig:gen_exp}
\end{figure}

\begin{table}[H]
\begin{center} 
\caption{Results of Gemma-2-9b-it based on knowledge injection, where “w/o"
denotes the performance without a specific module and \textit{kno\_emo} represents emotional knowledge.} 
\label{table2} 
\vskip 0.12in
\begin{tabular}{l c c c c} 
\hline
Method    &  Ratio    &  P(\%)    &  R(\%)    &  F1(\%) \\
\hline
MEKiT     &   1:1    &   63.73    &   56.74   &   60.03\\
                     &   1:2    &   \underline{64.44}    &   \underline{57.37}   &   \underline{60.70}\\
                     &   1:5    &   \textbf{65.04}    &   \textbf{58.31}   &   \textbf{61.49}\\
                     &   1:10    &   62.24    &   55.80   &   58.84\\
\midrule
MEKiT w/o kno\_emo                     &   1:1    &   62.72    &   \underline{56.43}   &   \underline{59.41}\\
                     &   1:2    &   61.72    &   56.11   &   58.78\\
                     &   1:5    &   \textbf{63.50}    &   \textbf{56.74}   &   \textbf{59.93}\\
                     &   1:10    &   \underline{62.81}    &   56.11   &   59.27\\
\hline
\end{tabular} 
\end{center} 
\end{table}

% \begin{figure}[t]
%   \includegraphics[width=\columnwidth]{img/general_exp_3.pdf}
%   \caption{Experimental results of different LLMs based on MEKiT.}
%   \label{fig:fig4}
% \end{figure}

\subsubsection{Generality of MEKiT}

We extend the MEKiT approach to other LLMs, including Vicuna-7B and LLaMA2-7B, to comprehensively evaluate its applicability and effectiveness. As shown in Figure~\ref{fig:gen_exp}, experimental results on different LLMs demonstrate that the lack of emotional knowledge and causal knowledge is a common phenomenon among such models and MEKiT significantly enhances model performance on the ECPE task, exhibiting high adaptability across various model architectures. Despite differences in parameter scales and pre-training strategies, the knowledge injection method consistently improves performance, indicating that our proposed framework possesses strong generality. Moreover, the performance of the Gemma-2-9B-it backbone consistently surpasses that of Vicuna-7B and LLaMA2-7B, likely due to its larger parameter scale, which enables it to more effectively capture complex semantic relationships and contextual nuances. In addition, we observe that the optimal mixing ratios for different LLMs are not necessarily consistent. For instance, Vicuna-7B achieves the best performance with the 1:10 ratio, while LLaMA2-7B performs best with the 1:2 ratio. This further confirms that excessively incorporating causal data does not necessarily lead to a continuous improvement in model performance.

% \begin{figure}[t]
%   \includegraphics[width=\columnwidth]{img/new_gen_exp.pdf}
%   \caption{Results of different LLMs based on MEKiT.}
%   \label{fig:gen_exp}
% \end{figure}

% \begin{table}[H]
% \begin{center} 
% \caption{Ablation study on the individual knowledge parts, where kno\_emo represents emotional knowledge and kno\_causal represents causal knowledge.} 
% \label{sample-table} 
% \vskip 0.12in
% \begin{tabular}{l c c c} 
% \hline
% Method    &  P(\%)    &  R(\%)    &  F1(\%) \\
% \hline
% MEKiT        &   \textbf{64.88}    &   \textbf{60.82}   &   \textbf{62.78}\\
% w/o kno\_emo + kno\_causal        &   61.05    &   54.55   &   57.62\\
% w/o kno\_emo        &   \underline{63.89}    &   \underline{57.68}   &   \underline{60.63}\\
% w/o kno\_causal        &   61.78    &   56.74   &   59.15\\

% \hline
% \end{tabular} 
% \end{center} 
% \end{table}

\begin{table}[H]
\begin{center} 
\caption{Ablation study on the individual knowledge parts, where \textit{kno\_emo} represents emotional knowledge and \textit{kno\_causal} represents causal knowledge.} 
\label{table3} 
\vskip 0.12in
\begin{tabular}{l c c c} 
\hline
Method    &  P(\%)    &  R(\%)    &  F1(\%) \\
\hline
MEKiT        &   \textbf{65.04}    &   \textbf{58.31}   &   \textbf{61.49}\\
w/o kno\_emo + kno\_causal        &   61.05    &   54.55   &   57.62\\
w/o kno\_emo        &   \underline{63.50}    &   \underline{56.74}   &   \underline{59.93}\\
w/o kno\_causal        &   61.78    &   \underline{56.74}   &   59.15\\

\hline
\end{tabular} 
\end{center} 
\end{table}

\subsection{Ablation Study}

We conduct an ablation study to analyze the impact of different components of multi-source heterogeneous knowledge, as shown in Table~\ref{table3}. The results indicate that the absence of any type of knowledge reduces in performance. Specifically, the removal of emotional knowledge and causal knowledge results in F1 score drops of 1.56\% and 2.34\%, respectively, with the most severe decline of 3.87\% when both are absent. These findings underscore the effectiveness of each component. We posit the main reason of result is that emotional knowledge enhances the model's emotional perception, enabling a more comprehensive understanding of emotional states, while causal knowledge improves the model's overall reasoning capability. We observe that the performance degradation is more pronounced without causal knowledge. The primary reason is that causal knowledge encompasses many contextual scenarios and explicit causal relationships that are absent in the original emotion-cause dataset, thereby introducing substantial additional information. According to the appraisal theory of emotion, the generation of emotion depends on cognitive evaluation of the situation and stimulus (the cause of the emotion), which is highly dependent on prior knowledge and experience. The enhancement of causal knowledge strengthens this background knowledge. However, emotional knowledge is derived through similarity calculation or polarity judgment based on commonsense knowledge generated from the dataset. Due to information overlap, only limited auxiliary knowledge can be introduced.

\subsection{Dicussion on Emotional Knowledge}

We conduct experiments on emotional knowledge generation tools utilizing emotion-english-distilroberta-base\footnote{https://huggingface.co/j-hartmann/emotion-english-distilroberta-base} (distilroberta), bert-base-cased-goemotions-original\footnote{https://huggingface.co/monologg/bert-base-cased-goemotions-original} (bert-goemotions) and bart-large-mnli\footnote{https://huggingface.co/facebook/bart-large-mnli} (bart-large-mnli), all pretrained for sentiment analysis. Both distilroberta and bert-goemotions output a single emotion category along with its corresponding score, while bart-large-mnli aligns with the format of our emotional label distribution knowledge. However, as shown in Table~\ref{table4}, integrating emotional knowledge generated by these models slightly decreases overall performance. We attribute this decline to the lower accuracy of fine-grained emotion classification compared to polarity classification, which introduces more noise and disrupts reasoning. In contrast, our method utilizing COMET can generate more context-specific information, better aligning with the contextual semantics. The label distribution knowledge we adopt is derived by calculating similarity scores based on commonsense context generated by COMET, followed by ranking. Furthermore, when commonsense context is ineffective, we introduce polarity knowledge, which exhibits higher accuracy compared to seven-category emotion classification. This method is more reliable than relying solely on single-category classification or ranking all emotion categories using pretrained models. Thus, in our research, we adopt a combined approach that leverages both methods to enhance reliability and performance.
\begin{table}[H]
\begin{center} 
\caption{Results of different emotional knowledge generation tools.} 
\label{table4} 
\vskip 0.12in
\begin{tabular}{l c c c} 
\hline
Method    &  P(\%)    &  R(\%)    &  F1(\%) \\
\hline
Gemma-2-9B-it        &   61.05    &   54.55   &   57.62\\
+kno\_distilroberta  &   60.35    &   54.86   &   57.47\\
+kno\_bert-goemotions  &   60.92    &   54.23   &   57.37\\
+kno\_bart-large-mnli  &   60.56    &   54.86   &   57.57\\
+kno\_label  &   \textbf{61.88}    &   \underline{55.49}   &   \underline{58.51}\\
+kno\_polarity  &   61.53    &   55.17   &   58.18\\
+kno\_label+kno\_polarity       &   \underline{61.78}    &   \textbf{56.74}   &   \textbf{59.15}\\

\hline
\end{tabular} 
\end{center} 
\end{table}

% \begin{table}[H]
% \begin{center} 
% \caption{Results of different emotion knowledge generation tools.} 
% \label{sample-table} 
% \vskip 0.12in
% \begin{tabular}{l c c c} 
% \hline
% Method    &  P(\%)    &  R(\%)    &  F1(\%) \\
% \hline
% Gemma-2-9b-it        &   61.05    &   54.55   &   57.62\\
% Gemma-2+DistilRoBERTa\_kno  &   60.35    &   54.86   &   57.47\\
% Gemma-2+BART\_kno  &   60.56    &   54.86   &   57.57\\
% Gemma-2+BERT\_kno  &   60.92    &   54.23   &   57.37\\
% Gemma-2+label\_kno  &   \textbf{61.88}    &   \underline{55.49}   &   \underline{58.51}\\
% Gemma-2+polarity\_kno  &   61.53    &   55.17   &   58.18\\
% Gemma-2+label\_kno+polarity\_kno       &   \underline{61.78}    &   \textbf{56.74}   &   \textbf{59.15}\\

% \hline
% \end{tabular} 
% \end{center} 
% \end{table}

\section{Conclusion}

In this paper, we propose a multi-source heterogeneous knowledge injection method MEKiT based on instruction-tuning, designed to address the ECPE task using the emotion cognitive appraisal theory. This approach compensates for the limited reasoning ability of small-scale pre-trained language models and tackles the challenges faced by LLMs in specific tasks, despite their strong general capabilities. Through experiments, we demonstrate that injecting multi-source knowledge, including emotional and causal knowledge, and applying instruction-tuning effectively enhance the performance of LLMs on the ECPE task, outperforming baselines. Overall, our research advances LLM capabilities in emotion cause reasoning.

\section{Acknowledgments}

This work is supported by the National Natural Science Foundation of China (62272092, 62172086) and the Fundamental Research Funds for the Central Universities of China (No. N2116008).

% \section{References Instructions}

% Follow the APA Publication Manual for citation format, both within the
% text and in the reference list, with the following exceptions: (a) do
% not cite the page numbers of any book, including chapters in edited
% volumes; (b) use the same format for unpublished references as for
% published ones. Alphabetize references by the surnames of the authors,
% with single author entries preceding multiple author entries. Order
% references by the same authors by the year of publication, with the
% earliest first.

% Use a first level section heading, ``{\bf References}'', as shown
% below. Use a hanging indent style, with the first line of the
% reference flush against the left margin and subsequent lines indented
% by 1/8~inch. Below are example references for a conference paper, book
% chapter, journal article, dissertation, book, technical report, and
% edited volume, respectively.

\nocite{ChalnickBillman1988a}
\nocite{Feigenbaum1963a}
\nocite{Hill1983a}
\nocite{OhlssonLangley1985a}
% \nocite{Lewis1978a}
\nocite{Matlock2001}
\nocite{NewellSimon1972a}
\nocite{ShragerLangley1990a}

\bibliographystyle{apacite}

\setlength{\bibleftmargin}{.125in}
\setlength{\bibindent}{-\bibleftmargin}

\bibliography{CogSci_Template}

\end{document}